\documentclass[conference]{IEEEtran}
\usepackage{times}

\usepackage[numbers]{natbib}
\usepackage{multicol}
\usepackage[bookmarks=true]{hyperref}

\usepackage{graphics} 
\usepackage{epsfig} 
\usepackage{mathptmx} 
\usepackage{amsmath} 
\usepackage{amssymb}  
\usepackage{units}

\usepackage{algorithm}
\usepackage[noend]{algpseudocode}
\usepackage{caption}
\usepackage{subcaption}
\usepackage{float}
\usepackage{multirow}
\usepackage{xspace}
\usepackage{wrapfig}

\DeclareMathOperator*{\argmax}{arg\,max}
\newcommand{\fdm}{FDM\xspace}
\newcommand{\its}{ITS\xspace}
\newcommand{\dtw}{DTW\xspace}

\newcommand{\approximate}{Approximate\xspace}
\newcommand{\complete}{Complete\xspace}
\newcommand{\pd}{PD\xspace}
\newcommand{\squeezeup}{\vspace{-5mm}}

\begin{document}

\title{Learning Forward Dynamics Model and Informed \\
Trajectory Sampler for Safe Quadruped Navigation}

\author{\authorblockN{Yunho Kim, Chanyoung Kim, Jemin Hwangbo}
\authorblockA{Korea Advanced Institute of Science and Technology (KAIST), Republic of Korea\\ \{awesomericky, slowturtle99, jhwangbo\}@kaist.ac.kr}}

\maketitle

\begin{abstract}
For autonomous quadruped robot navigation in various complex environments, a typical SOTA system is composed of four main modules -- mapper, global planner, local planner, and command-tracking controller -- in a hierarchical manner. In this paper, we build a robust and safe local planner which is designed to generate a velocity plan to track a coarsely planned path from the global planner. Previous works used waypoint-based methods (e.g. Proportional-Differential control and pure pursuit) which simplify the path tracking problem to local point-goal navigation. However, they suffer from frequent collisions in geometrically complex and narrow environments because of two reasons; the global planner uses a coarse and inaccurate model and the local planner is unable to track the global plan sufficiently well. Currently, deep learning methods are an appealing alternative because they can learn safety and path feasibility from experience more accurately. However, existing deep learning methods are not capable of planning for a long horizon. In this work, we propose a learning-based fully autonomous navigation framework composed of three innovative elements: a learned forward dynamics model (\fdm), an online sampling-based model-predictive controller, and an informed trajectory sampler (\its). Using our framework, a quadruped robot can autonomously navigate in various complex environments without a collision and generate a smoother command plan compared to the baseline method. Furthermore, our method can reactively handle unexpected obstacles on the planned path and avoid them. (\href{https://awesomericky.github.io/projects/FDM_ITS_navigation/}{Video\footnote{Supplementary materials: \href{https://awesomericky.github.io/projects/FDM_ITS_navigation/}{\\ awesomericky.github.io/projects/FDM\_ITS\_navigation/}}})
\end{abstract}

\IEEEpeerreviewmaketitle


\section{Introduction}

Thanks to recent advances in legged robot control research~\cite{hwangbo2019learning, bellicoso2018dynamic, gong2019feedback, xie2018feedback, haarnoja2018learning}, legged robots have become more robust and more agile in diverse environments. Notably, learning-based control approaches have shown great performance in complex outdoor environments. Unfortunately, those are largely blind: they only use proprioceptive sensors~\cite{lee2020learning} or very local information of the terrain~\cite{miki2022learning}. Furthermore, they can follow the given velocity commands but it is not trivial how we can generate adequate velocity commands in various environments. For autonomous navigation, the robot should be able to plan a safe and efficient plan.

We are interested in a \textit{point-goal navigation} task, where the legged robot should autonomously navigate to the given goal position using real-time exteroceptive and proprioceptive sensor data. For robust autonomous navigation, many previous works ~\cite{kim2020vision, dudzik2020robust, chilian2009stereo, wermelinger2016navigation, hildebrandt2017real, gilroy2021autonomous, li2021vision, wooden2010autonomous} proposed a hierarchical framework composed of four main modules: mapper, global planner, local planner, and command-tracking controller. The roles of each modules are as following:

\begin{itemize}
    \item Mapper: Build a map of the environment using exteroceptive sensors such as a lidar and camera.
    \item Global planner: Find a rough path from the current location to the goal location using the map built by the mapper. A variety of planning algorithms such as search-based methods (e.g. A*~\cite{hart1968formal}), sampling-based methods (e.g. RRT~\cite{lavalle1998rapidly}, PRM~\cite{latombe1998probabilistic}, RRT*, PRM*~\cite{karaman2011sampling}), and heuristic methods (e.g. Potential Field~\cite{khatib1986real}) can be used.
    \item Local planner: Generate a high-level command for the robot to track the planned path from the global planner. During the process, the roughly planned path is projected to a locally feasible trajectory that the robot can track. Various types of high-level commands can be planned by the local planner. In this work, we will focus on a local planner generating velocity commands.
    \item Command-tracking controller: Generate a joint-level command to track the given high-level command from the local planner. A learning-based~\cite{hwangbo2019learning} or model-based~\cite{bellicoso2018dynamic} controller can be used for this purpose.
\end{itemize}

\begin{figure}[t]
    \centering
    \includegraphics[width=\linewidth]{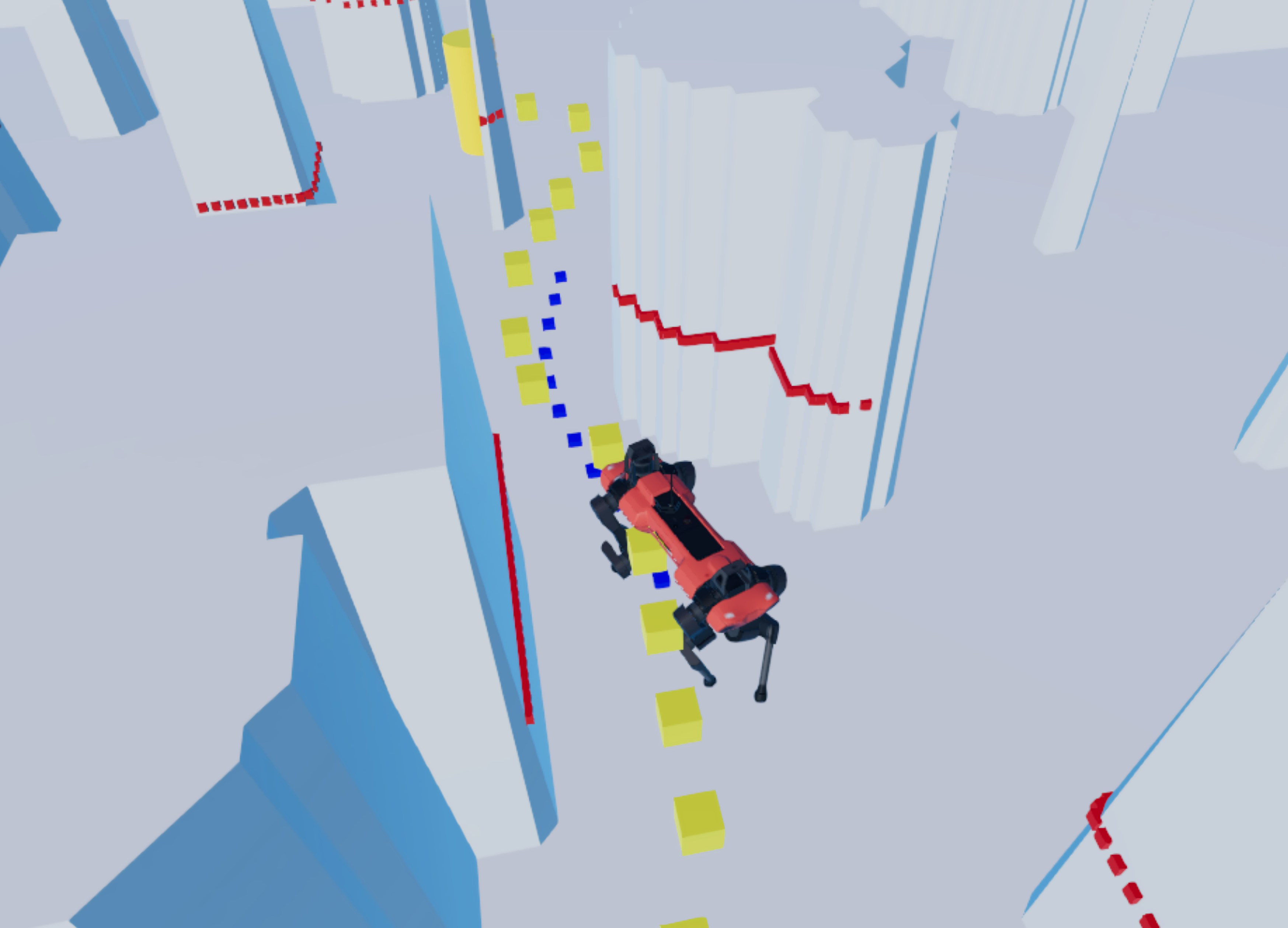}
    \caption{Using our safe navigation framework, quadruped robot can safely navigate in complex environments. The yellow cylinder is the goal position. The yellow dotted line is the planned path by the global planner. The blue dotted line is the generated trajectory by the optimized command sequence.}
    \label{fig:eyecatching}
    \squeezeup
\end{figure}

All four modules should work in concert for safe and robust navigation. However, compared to recent advances in mapping and global planning techniques~\cite{hess2016real, Fankhauser2018ProbabilisticTerrainMapping, gammell2015batch, gammell2014informed}, there were relatively little developments in local planning. In geometrically complex environments, local planning becomes a challenging problem. In these environments, the local planner should be capable of generating long horizon plans that satisfies the kinematic and dynamics constraints of the robot and the performance characteristics of the command-tracking controller. Furthermore, it has to be reactive enough to handle unexpected obstacles that have not been accounted for during global planning.

In this light, we propose a learning-based, fully autonomous navigation framework composed of three innovative elements: a learned forward dynamics model (\fdm), an online sampling-based model-predictive control module, and an informed trajectory sampler (\its). We demonstrate how the dynamics of a quadruped robot and its surroundings can be captured accurately using deep neural networks (i.e. \fdm), which is trained with a self-supervised learning framework, and used for predicting the future outcomes of a given command sequence. We then used \fdm in conjunction with an online sampling-based model-predictive control module to track a roughly planned path from the global planner, as illustrated in Fig. \ref{fig:eyecatching}. To handle the curse of dimensionality in sampling-based motion planning, we learned the implicit command sampling distribution (i.e. \its) which generates samples close to optimal solutions and used it with the proposed online sampling-based model-predictive control module. 

Because the learned \fdm can be evaluated very fast in a GPU-enabled hardware, we can simulate 1,500 of 6-second trajectories in a \unit[3]{ms} window, which is more than 20,000 times and 20 times faster than the simulation of the full and approximate robot model, respectively. An extensive evaluation in a physics simulator~\cite{hwangbo2018per} shows that the proposed method for local planning enables safer autonomous navigation in diverse geometrically complex environments compared to widely used baseline methods. Furthermore, the proposed method can reactively handle unexpected obstacles on the planned path and be easily modified for both fully-autonomous and semi-autonomous tasks.


\section{Related work} \label{related_work}

To generate a velocity command to track a path from the global planner, many previous works~\cite{dudzik2020robust, chilian2009stereo, wermelinger2016navigation, wooden2010autonomous} used waypoint-based methods which simplify the path tracking problem into local point-goal navigation. These methods repeat (1) determining a waypoint on the planned path that is within an adequate distance range and (2) generating a command to move toward the determined waypoint. Commands are generated using a PD (i.e. Proportional-Differential) controller based on the position and orientation error between the current and desired configuration in the waypoint. However, these methods suffer from frequent collisions in complex and narrow environments. Enlarging the approximated robot collision body can be a solution for the global planner to find a more conservative path, but it results in a compromising path or failure to find one. Post-processing the planned path heuristically~\cite{wooden2010autonomous} for smoothness still cannot be a complete solution to guarantee safety in local planning. Some works used reactive methods~\cite{mattamala2022efficient, kim2020vision, hildebrandt2017real} with vector field representation for local obstacle avoidance, but they result in jerky and discrete command changes due to the lack of long horizon planning. ~\citet{gilroy2021autonomous} and ~\citet{li2021vision} used collocation-based trajectory optimization to handle long horizon planning, but the decoupled nature between the local planner and the command-tracking controller can still cause a collision in complex environments because the local planner cannot take into account the performance characteristics of the command-tracking controller. Furthermore, these methods cannot handle unexpected obstacles on the globally planned path.

Learning-based methods are an appealing alternative because they can accurately learn the safety and path feasibility from experience. Previous works~\cite{tai2016deep, tai2017virtual, pfeiffer2017perception, hoeller2021learning} used imitation learning and model-free deep reinforcement learning to learn a collision-free control policy for a wheeled mobile robot and a quadruped robot. However, the results were a reactive single-step planning policy and also cannot be integrated with the existing hierarchical navigation framework. When deploying the resulting policy for a \textit{point-goal navigation} task, the robot will easily get stuck in a local optimum due to the lack of a long horizon planning module.

There were recent advances in the global planner using learning-based methods to find the global path faster~\cite{bhardwaj2017learning, qureshi2018deeply, qureshi2019motion, ichter2018learning, zhang2018learning} or to find a controller-aware path~\cite{guzzi2020path, yang2021real}. However, these methods still require a robust local planner to safely track the planned path.

Our work is closely related to the one presented by Kahn et al.~\cite{kahn2021badgr}. They presented a forward dynamics model, composed of deep neural networks, that predicts the path that the robot will take, the corresponding collision probabilities, and the terrain properties. Our work differs from theirs in three ways. First, their focus was on identifying the traversability of the terrain from an RGB image in open field environments and use this information in finding a command trajectory that leads the robot toward the goal. Because the environments they experimented had less obstacles, the robot can safely navigate to the goal without a globally planned path. However, in geometrically complex environments that we are interested in, the robot will easily get stuck in a local optimum without using the global planner. Thus, we focused on finding a command trajectory to track the planned path from the global planner in environments with densely placed obstacles, requiring a more reactive controller and a different problem formulation. Second, we improved the sampling-based motion planner used by ~\citet{kahn2021badgr} by learning the implicit command sampling distribution (i.e. \its) using Conditional Variational Inference~\cite{sohn2015learning}, which results in a safer motion plan. Lastly, we tested our controller with a legged robot, which manifests more complicated dynamics and kinematics compared to wheeled mobile robots, and a 360-degree lidar sensor readings for safe navigation in all directions, rather than front view camera images.


\section{Method}

\begin{figure}[t]
    \centering
    \includegraphics[width=\linewidth]{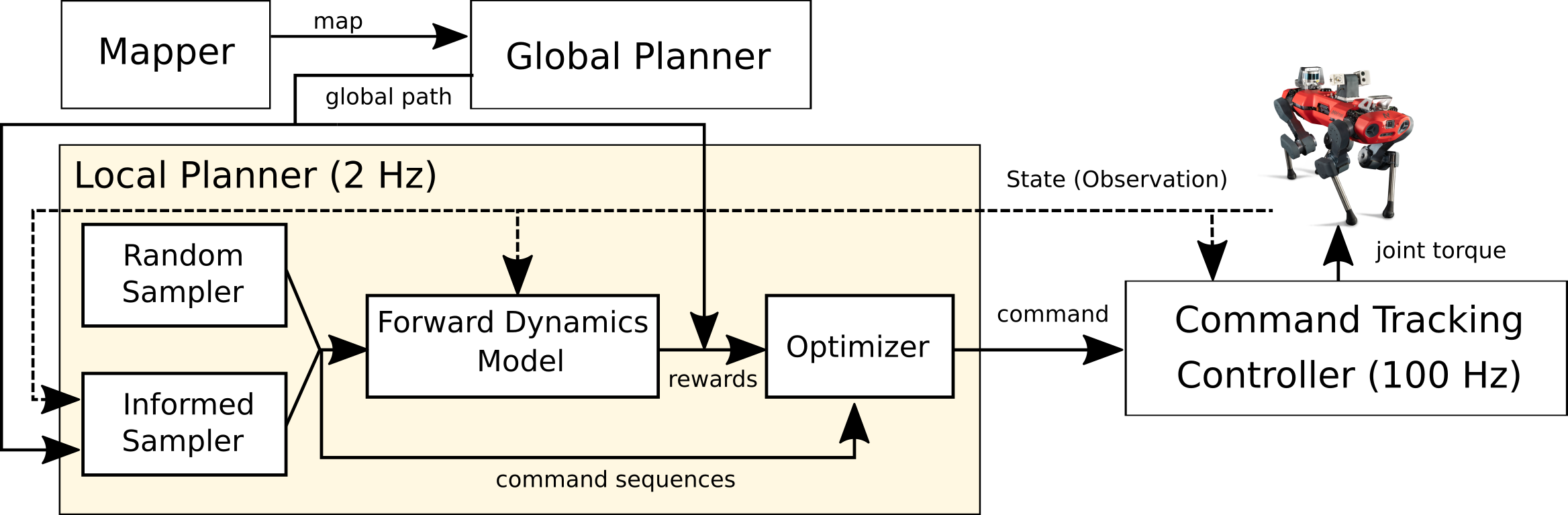}
    \caption{Block diagram of the proposed safe navigation framework.}
    \label{fig:navigation_framework}
    \squeezeup
\end{figure}

Our goal is to improve the existing hierarchical navigation system by building a more robust and safe local planner that can generate a sequence of velocity commands to track the coarsely planned path from the global planner in a geometrically complex environment with flat terrain and many static obstacles. Thus, the proposed safe navigation framework is constructed similarly to the existing navigation system with four main modules: mapper, global planner, local planner, and command-tracking controller (Fig. \ref{fig:navigation_framework}). Although our framework can be easily combined with a mapper for navigation in unknown environments similar to previous works~\cite{li2021vision, gilroy2021autonomous}, we assume that the map is given ahead of time, because it is not the focus of this paper. For global planning, we used an open-source implementation~\cite{sucan2012the-open-motion-planning-library} of BIT*~\cite{gammell2015batch} because of its speed and robustness. However, many other path planning algorithms~\cite{hart1968formal, karaman2011sampling, gammell2014informed} can be used with our framework as well. The global planner finds a rough path from the start location to the goal location and returns a list of x, y coordinates, represented in the world frame, to the local planner. The global planner does not consider the orientation of the robot due to the computation complexity in real-time usage and uses a minimal bounding sphere to simplify the robot's collision bodies~\cite{wermelinger2016navigation}.

The local planner is composed of three core elements: a learned forward dynamics model (\fdm), an online sampling-based model-predictive control module, and an informed trajectory sampler (\its) (Fig. \ref{fig:navigation_framework}). In the rest of the subsections, we will introduce each element and summarize the overall working pipeline. For the command-tracking controller, we use a controller proposed in our previous work, which is acquired using model-free deep reinforcement learning~\cite{hwangbo2019learning}.

We use a simulated ANYmal C robot~\cite{hutter2016anymal} for both training and testing of our proposed framework in the RaiSim simulator \cite{hwangbo2018per}. For perception, we use a 2D line-scan lidar sensor attached to the back of the robot. Lidar readings are modeled with a Gaussian noise \textit{N}(0, 0.2) [m] and normalized with the maximum available sensing range of 10 m.

\begin{figure*}[t]
    \centering
    \includegraphics[width=\linewidth]{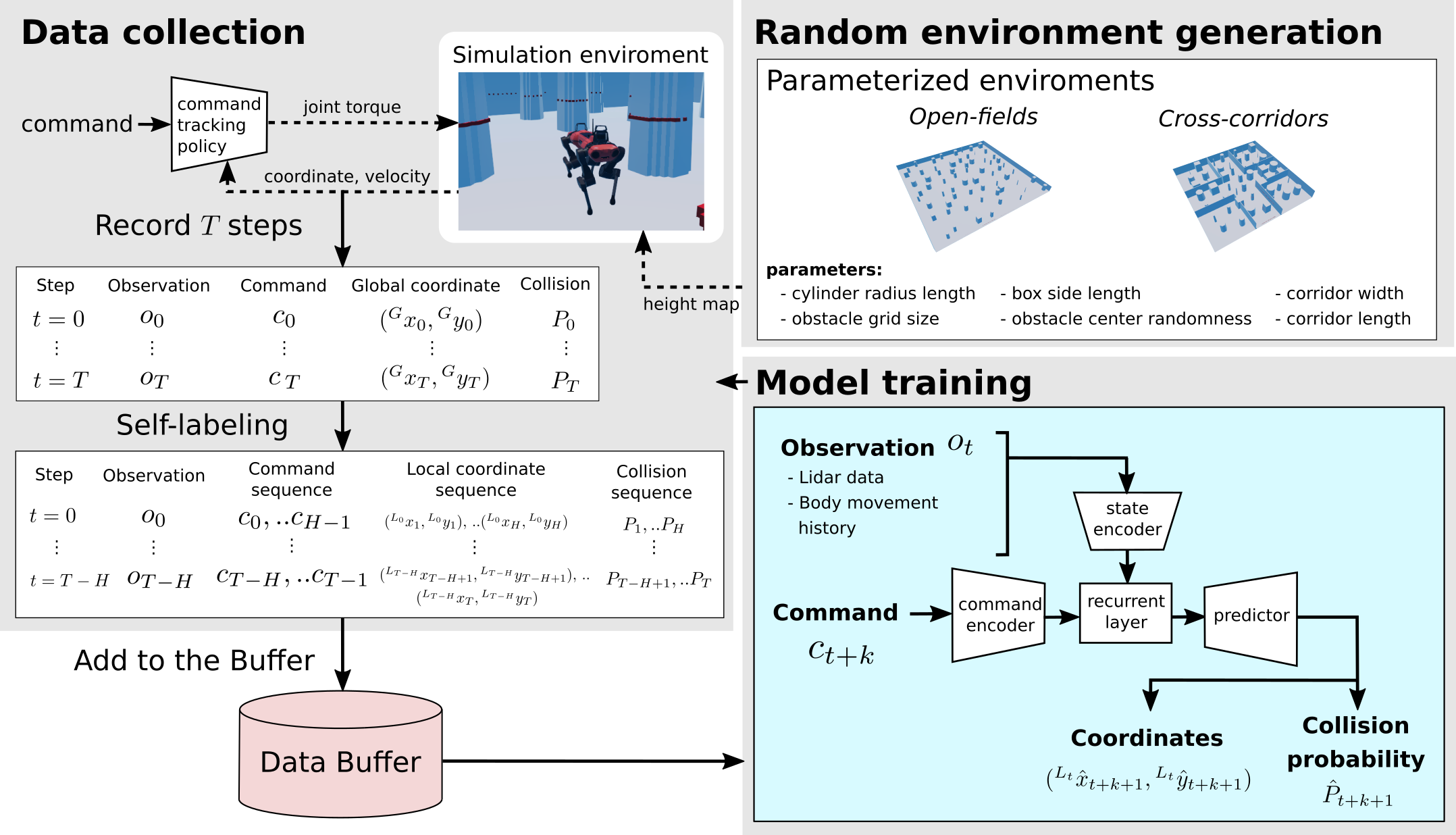}
    \caption{Overall framework to train \fdm. The neural network architecture of \fdm was inspired by the ones used by ~\citet{kahn2021badgr}.}
    \label{fig:fdm_training_framework}
    \squeezeup
\end{figure*}

\subsection{Forward dynamics model}

To plan for a long horizon, we learn the dynamics of the environment rather than a task-specific control policy, inspired by numerous previous works~\cite{finn2017deep, nagabandi2020deep, kahn2021badgr, kahn2018self}. Our learned \fdm works as a fast virtual simulator for a quadruped robot and predicts the future base coordinates and the probability of collision on the trajectory. To be specific, it takes as input the current lidar sensor data, history of selected generalized coordinates and velocities, and a sequence of future intended commands, and predicts the future x, y coordinates of the robot with respect to the robot's current body frame and the probability of collision. The history of selected generalized coordinates and velocities correspond to the history of base orientation, base linear velocity, and base angular velocity (10 steps of history, each step corresponding to 0.05 second). The history term was included due to the dynamic movements of the origin of the lidar frame. We denote this model as $f_{\theta}(\textbf{o}_t, \textbf{c}_{t:t+H}) = \{\hat{\textbf{x}}_{t+1:t+H+1}, \hat{\textbf{y}}_{t+1:t+H+1}, \hat{\textbf{p}}_{t+1:t+H+1}\}$, which is parameterized by the vector $\theta$, that takes as input the current observation $\textbf{o}_t$ and a sequence of $H$ future commands $\textbf{c}_{t:t+H}=\{\textbf{c}_t, \textbf{c}_{t+1}, ..., \textbf{c}_{t+H-1}\}$. The model predicts the x, y coordinates $\hat{\textbf{x}},\ \hat{\textbf{y}}$ and probability of collision $\hat{\textbf{p}}$ for $H$ future time steps.

\fdm is a deep neural network composed of fully connected layers and subsequent recurrent layers to predict the future positions and collision probabilities conditioned on the current observations and a sequence of commands. The current observations, which include both proprioceptive and exteroceptive sensor information, are encoded to latent features using fully connected layers. The final output of these layers serves as a hidden state initialization for a recurrent neural network, which sequentially processes each of the $H$ future commands $\textbf{c}_{t:t+H}$ and outputs the future navigational outcomes. We use the LSTM (i.e. Long Short-Term Memory) cells for the recurrent layers \cite{hochreiter1997long}.

To let \fdm learn the dynamics of the environment, we focus on generating diverse environments to capture both a broad lidar data distribution and a future outcome distribution. We train \fdm on two types of parameterized environments: open-fields with densely placed cylinders/boxes, and cross-shaped corridors with densely placed cylinders/boxes. Each of these environments is randomly generated by sampling the corresponding parameters from the predefined range given in Table \ref{tab:env_generation}. The map is divided into equal sizes of grids, and each grid contains one obstacle. The position of the obstacle inside the grid is sampled from $U(center\ randomness,\ grid\ size - center\ randomness)$, where $center\ randomness$ determines the overall randomness of the obstacle distribution. Using our parameterization, we can generate various environments with both convex and concave obstacles, due to the occlusion between cylinders/boxes, and let \fdm learn the general ability to detect collisions.

\begin{table}[t]
    \centering
    \begin{tabular}{l|c|c}
    \hline Environment type & Parameter & Sampling distribution \\
    \hline
    \hline \multirow{4}{*}{Open-fields} & cylinder radius & \textit{U}(0.05, 1.0) [m] \\
    \cline{2-3}                        & box side        & \textit{U}(0.1, 2.0) [m] \\
    \cline{2-3}                        & grid size       & \textit{U}(2.3, 5.0) [m] \\
    \cline{2-3}                        & center randomness & \textit{U}(0.1, 0.9) [m] \\
    \hline \multirow{6}{*}{Cross-corridors} & cylinder radius & \textit{U}(0.05, 1.0) [m] \\
    \cline{2-3}                             & box side        & \textit{U}(0.1, 2.0) [m] \\
    \cline{2-3}                             & grid size       & \textit{U}(2.3, 5.0) [m] \\
    \cline{2-3}                             & center randomness & \textit{U}(0.1, 0.9) [m] \\
    \cline{2-3}                             & corridor width    & \textit{U}(2.0, 6.0) [m] \\
    \cline{2-3}                             & corridor length   & \textit{U}(8.0, 30.0) [m] \\
    \hline
    \end{tabular}
    \caption{Parameters for the random environment generation}
    \label{tab:env_generation}
    \squeezeup
\end{table}

To take advantage of the fast parallel data generation process of the simulator, we alternate between a data collection and model training step. In the data collection step, we randomly generate $N_{env}$ environments with different types of obstacles in the same proportion. During the data collection period, a quadruped robot is randomly placed in each environments and given velocity command sequences sampled from \{\textit{U}(-1.0, 1.0), \textit{U}(-0.4, 0.4), \textit{U}(-1.2, 1.2)\}, each elements corresponds to forward velocity [m/s], lateral velocity [m/s], and turning rate [rad/s]. The specific command sequence sampling method used for training \fdm is described in APPENDIX-A.

We use a self-supervised data labeling technique, similar to \cite{kahn2021badgr}, to process the data for training without human supervision. Simple coordinate transformation is used to compute x, y coordinates in the robot's current body frame. For the probability of collision, binary collision state, computed by solving the contact dynamic in the physics simulation, is used \cite{hwangbo2018per}. Data tuples $\{\textbf{o}_t, \textbf{c}_{t:t+H}, \textbf{x}_{t+1:t+H+1}, \textbf{y}_{t+1:t+H+1}, \\ \textbf{p}_{t+1:t+H+1}\}$ are recorded to the data buffer and used later for training.

The model is trained to minimize a loss function that penalizes the distance between the predicted and actual outcomes. The mean squared error and cross-entropy loss are used for the x, y coordinates and probability of collision, respectively. The overall training framework is summarized in Fig. \ref{fig:fdm_training_framework}. Hyperparameters used for the data collection and training are shown in the APPENDIX-B.

We train \fdm to predict 12 steps of future navigational outcomes, each step corresponding to \unit[0.5]{s} period which results in \unit[6]{s} of the maximum prediction horizon, and use it for the experiments done in section \ref{experiment}.

\subsection{Online sampling-based model-predictive control}

In this section, we present an online sampling-based model predictive control algorithm using learned \fdm, which works as a fast virtual simulator with collision checking, for safe quadruped navigation. We define a reward function $R(f_{\theta}(\textbf{o}_t, \textbf{c}_{t:t+H}))$ in terms of the future navigational outcomes predicted by \fdm, conditioned on the given task of the robot. Using this reward function and \fdm, we repeatedly solve the following optimization problem,
\begin{equation} \label{mpc_objective}
    \textbf{c}_{t:t+H}^{*} = \argmax_{\textbf{c}_{t:t+H} \in \mathcal{C}}  R(f_{\theta}(\textbf{o}_t, \textbf{c}_{t:t+H})), 
\end{equation}
and execute the first step of the resulting command sequence for a single command period, where $\mathcal{C}$ is the available command range.

To solve Eqn. \ref{mpc_objective}, we use a gradient-free optimizer similar to the ones used by Nagabandi et al.~\cite{nagabandi2020deep} and Kahn et al.~\cite{kahn2021badgr}. At every time step, $N$ commands $\mathit{c}_{t:t+H}^{0:N}$ are generated by computing the weighted average of time-correlated random command sequences $\tilde{\mathit{c}}_{t:t+H}^{0:N}$ and previously optimized results $\hat{\textbf{c}}_{t:t+H}$ as 
\begin{equation} \label{mpc_command_sampling}
\begin{split}
    \mathit{c}_{t:t+H}^n\ =\ (1-\beta) \cdot \tilde{\mathit{c}}_{t:t+H}^n\ +\ \beta \cdot \hat{\textbf{c}}_{t:t+H} \\
    \text{s.t.} \quad \tilde{\mathit{c}}_{t+k+1}^{n}\ \sim\ \textit{N} (\tilde{\mathit{c}}_{t+k}^{n}, \sigma) \\
    \tilde{\mathit{c}}_{t}^{n}\ \sim\ Bin(\mathit{c}_{min}, \mathit{c}_{max}) \\
    \forall\ k \in \{0..,H-2\},\ \forall\ n \in \{0..,N-1 \} \\
\end{split}
\end{equation}
where $Bin$ is a bin sampling method, which divides the command range into $N_b$ equally spaced bins and samples uniformly from each of them.

Each command sequence is then fed to \fdm to compute the corresponding reward $ R^n\ =\ R(f_{\theta}(\textbf{o}_t, \mathit{c}_{t:t+H}^n))$. Using the sampled command sequences and computed rewards, we update the optimized command sequence via reward-weighted average as 
\begin{equation} \label{mpc_command_update}
    \hat{\textbf{c}}_{t:t+H}\ =\ \frac{\sum_{n=0}^{N-1} \exp(\gamma \cdot R^n) \cdot \mathit{c}_{t:t+H}^n}{\sum_{n'=0}^{N-1} \exp(\gamma \cdot R^{n'})},
\end{equation}
similar to the equation by recent model-predictive path integral work \cite{williams2015model, lowrey2018plan}.

We then execute the first step of the planned command sequence for a single command period and repeat the process. Parameters $\beta \in [0, 1]$ and $\gamma \in \mathrm{R}^+$ of the optimizer work as a time correlation factor and a high-reward weighting factor, respectively. The optimizer we used shows stable performance in complex environments due to the ability to sample a dense set of trajectories. The performance of our algorithm will be further analyzed in section \ref{experiment}. Our overall algorithm is summarized as follows (Alg. \ref{alg:mpc}).

\begin{algorithm}[!h]
\caption{Online sampling-based model-predictive control with learned \fdm}
\label{alg:mpc}
\begin{algorithmic}[1]
\State \textbf{input}: learned \fdm $f_\theta$, reward function $R$
\While{task is not complete}
        \State get current observation $\textbf{o}_t$ from sensors
        \State solve Eqn. \ref{mpc_objective} using Eqn. \ref{mpc_command_sampling} and Eqn. \ref{mpc_command_update}
        \State execute $\hat{\textbf{c}}_{t}$ in optimized command sequence $\hat{\textbf{c}}_{t:t+H}$
\EndWhile
\end{algorithmic}
\end{algorithm}

\subsection{Implementation for path tracking}

For the robot to safely track the planned path, we use the sum of two rewards, $R_{track}$ and $R_{safety}$, defined as
\begin{equation}\label{path_track_reward}
\begin{split}
    &R_{total} = R_{track} + R_{safety} \\
    & s.t.\ R_{track} = exp(\frac{-DTW(\textbf{g}_t, \textbf{q}_{t+1:t+H+1})}{\tau}) \\
    & \quad \ R_{safety} = \frac{\sum_{k=1}^{H} (1 - \hat{\textbf{p}}_{t+k})}{H} \\
\end{split}
\end{equation}
where $\textbf{g}_t = {}^{L_t} \textbf{g}$ ($\textbf{g} \subset \textbf{G}$, $\textbf{G}$: Global path), $\textbf{q}_{t+k} = \{{}^{L_t} \hat{\textbf{x}}_{t+k}, {}^{L_t} \hat{\textbf{y}}_{t+k}\}$, and $\tau$ is a temperature constant for normalization.

$R_{track}$ is to incentivize command sequences that result in a similar path with the currently considered global path. Thus, we used the output of normalized Dynamic Time Warping (\dtw) for $R_{track}$. \dtw is a similarity function between two series of data by finding their optimal alignment which minimizes the cumulative distance between aligned elements. Because \dtw does not require two series of data to have the same length or constant difference in successive data, it has recently been used as an evaluation metric for navigation tasks~\cite{ilharco2019general}. In this work, we applied it in continuous planning for path tracking.

To compute $R_{track}$, we truncate the globally planned path that is 4.8m ahead from the current location, resulting in 0.8 m/s of desired average speed, and transform the coordinates of the considered path into the robot's local frame. \dtw between the considered path in the robot's local frame and the predicted path from \fdm is computed and normalized. Open-source implementation~\cite{giorgino2009computing} is used for computing \dtw. The truncated global path considered in the current step will be referred to as the waypoint trajectory hereinafter.

To generate safer navigation plan, the collision probability outputs from \fdm is used for both hard and soft constraints. For hard constraint, we filter out the sampled command trajectories that collide with obstacles within a fixed period (3s). For soft constraint, we compute $R_{safety}$ and include it in the total reward formulation to give additional weights to the trajectories that show a low probability of collision.

The planning module runs at 2Hz. In each planning period, we plan 12 steps of command trajectory to track the global path, each step corresponding to 0.5s period which results in 6s of the maximum planning horizon.

\subsection{Informed trajectory sampler}

Although the robot showed overall high performance by just sampling random time-correlated command sequences as Eqn. \ref{mpc_command_sampling}, it is still not free from the curse of dimensionality in sampling-based motion planning and thus often fails in complex environments. Considering that the command dimension is three, the number of planning steps is $H$, and the number of bins used for bin sampling is $N_b$, we need at least ${N_b}^{3H}$ number of samples to guarantee a near-optimal solution in any cases. However, as we are building methods that are capable of long-distance planning (i.e. large $H$) it is impossible to consider ${N_b}^{3H}$ samples online. 

Therefore, we additionally use the informed trajectory sampler (\its) that can generate command sequences close to the optimal solution. 

\its is a deep neural network modeled with Conditional Variational AutoEncoder (CVAE)~\cite{sohn2015learning}. CVAE is a conditional generative model that uses conditional variational inference to handle the intractable posterior distribution. The model is trained by optimizing the Evidence Lower BOund (ELBO) as shown in
\begin{equation} \label{cvae}
\begin{split}
    log p_{\theta}(x|y)\\ =\ \mathbb{E}_{z \sim q_{\phi}(z|x, y)}[log p_{\theta}(x|z, y)]\ -\ D_{KL}(q_{\phi}(z|x, y) || p_{\theta}(z|y)) \\ +\ D_{KL}(q_{\phi}(z|x, y) || p_{\theta}(z|x, y)) \\
    \geq\ \mathbb{E}_{z \sim q_{\phi}(z|x, y)}[log p_{\theta}(x|z, y)]\ -\ D_{KL}(q_{\phi}(z|x, y) || p_{\theta}(z|y)) \\
    =\ \underset{ELBO}{\underline{\mathbb{E}_{z \sim q_{\phi}(z|x, y)}[log p_{\theta}(x|z, y)]\ -\ D_{KL}(q_{\phi}(z|x, y) || p_{\theta}(z))}}. \\
    (\because \textit{Assume y and z independent}) \\
    (\textit{$D_{KL}$ corresponds to Kullback–Leibler divergence})
\end{split}
\end{equation}

In our case, $x$ corresponds to a command trajectory and $y$ corresponds to an observation and a global path considered in the current step (i.e. waypoint trajectory). Observation includes the current lidar sensor data and history of selected generalized coordinates and velocities, same as \fdm. Waypoint trajectory is represented as a list of x, y coordinates included in the globally planned path and transformed into the robot's local frame. 

Because sampling-based path planning algorithms, such as BIT*, output list of nodes that show irregular distances between them, waypoint trajectory sometimes include very few nodes. In this case, we interpolate it to guarantee a minimum number of nodes.

\its is composed of fully connected layers and recurrent layers. We use the GRU (i.e. Gated Recurrent Unit) cells for the recurrent layers \cite{cho2014learning} and sequence-to-sequence model to encode and decode command trajectory and waypoint trajectory ~\cite{sutskever2014sequence}. 160K number of training data is collected by rolling out the robot, controlled with Alg. \ref{alg:mpc}, in randomly generated environments (Fig. \ref{fig:fdm_training_framework}) and goal positions. For the learned sampler to generate locally diverse samples, we use a variant of the CVAE objective function proposed by ~\citet{bhattacharyya2018accurate}. Hyperparameters used for training \its are shown in the APPENDIX-B.

\begin{figure*}[t]
    \centering
    \includegraphics[width=\linewidth]{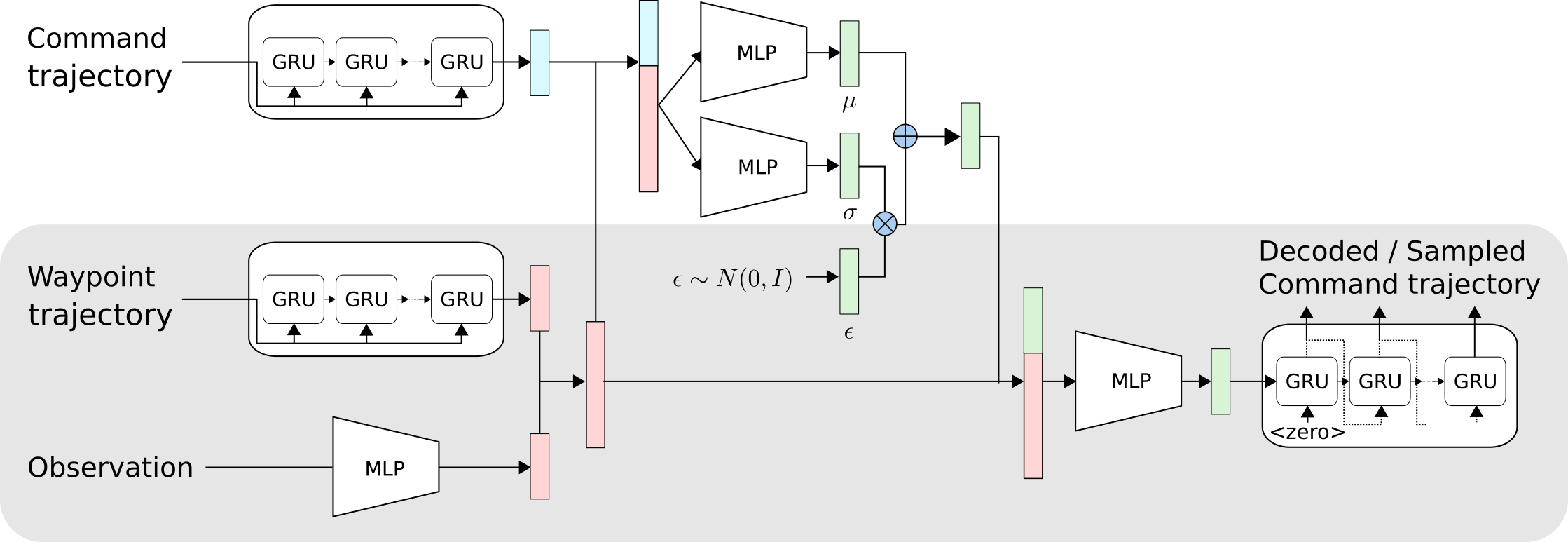}
    \caption{Model architecture to train \its. Gray scaled part is \its used for generating command trajectory samples in real-time navigation. The neural network architecture of \its was inspired by the ones used by ~\citet{ichter2018learning} and ~\citet{lee2017desire}. MLP in the figure indicates fully connected layers.}
    \label{fig:its_architecture}
\end{figure*}

Learned \its is then used with the time-correlated random sampler to generate command sequences for the online sampling-based model-predictive control module. The detailed model architecture of \its is shown in Fig. \ref{fig:its_architecture}.

\subsection{Summary}

We now provide a summary of our safe navigation framework (Fig. \ref{fig:navigation_framework}). $N$ command sequences are first sampled from both a time-correlated random sampler and \its. For each sample, the future navigation outcomes and rewards are then predicted with \fdm and Eqn. \ref{path_track_reward}. Based on Eqn. \ref{mpc_command_update}, the optimum command sequence is predicted and the first step command of it is executed by passing it to the command tracking controller, which outputs the target joint torque. This process repeats until the task is completed.


\section{Experiment} \label{experiment}

In our experiments, we study how the proposed learning-based safe navigation framework enables a quadruped robot to autonomously navigate in various complex environments. Hyperparameters of the sampling-based model-predictive controller were set constant for all the experiments (APPENDIX-B).

In the experiments, we used AMD Ryzen9 5950X and NVIDIA GeForce RTX 3070 for computation. \fdm and \its, which are composed of neural networks, inference were done on GPU (i.e. Graphics Processing Unit) and other computations such as sampling-based optimization were done on CPU (i.e. Central Processing Unit).

\subsection{\fdm evaluation} \label{fdm_evaluation}

We first evaluated the performance of \fdm in terms of prediction accuracy and computation speed. As \fdm works as a fast virtual simulator to predict the future navigation outcomes, the performance was compared with two baseline models that use a real-time collision checking algorithm in the physics simulator~\cite{hwangbo2018per}.
\begin{itemize}
    \item \textbf{\complete}: A method that runs the forward simulation of a quadruped robot with complete kinematic and dynamic models. Velocity tracking controller~\cite{hwangbo2019learning} was used to map the given velocity command to a joint-level command.
    \item \textbf{\approximate}: A method that runs the forward simulation of a quadruped robot with approximated kinematic models. For fast collision checking, the robot was approximated to a box with a size that covers the entire body of it in the nominal configuration. Future coordinates were computed analytically by assuming that the robot follows the velocity command perfectly.
\end{itemize}
For the baseline methods, we assumed the terrain model of the environment is known apriori for the physics simulator to run a collision checking algorithm.

\begin{figure*}[t]
    \centering
    \includegraphics[width=\linewidth]{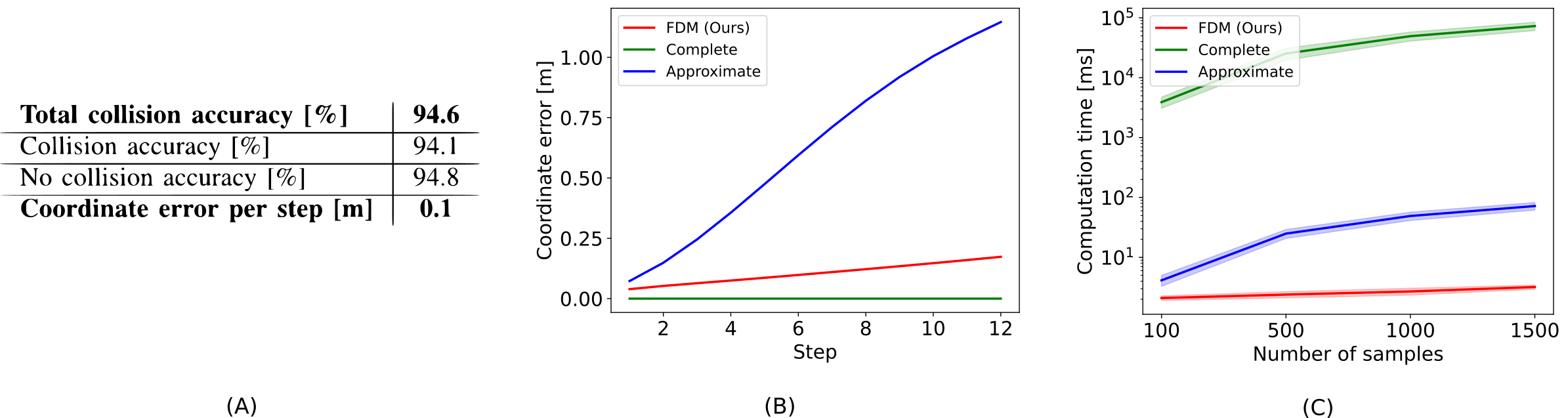}
    \caption{\fdm evaluation results. (A) shows the overall prediction accuracy of \fdm. (B) and (C) show the comparison between the baseline models in terms of coordinate prediction error and computation time.}
    \label{fig:fdm_evaluation}
\end{figure*}

The accuracy of \fdm was evaluated on 3.5M samples collected in randomly generated environments, which were not seen during training. The generated environments include both open-fields with densely placed cylinders/boxes, and cross-shaped corridors with densely placed cylinders/boxes (Fig. \ref{fig:fdm_training_framework}, Table \ref{tab:env_generation}). A probability threshold of 0.3 was used for deciding collision.

\fdm showed high collision checking accuracy (94.6\%) and low coordinate prediction error per step (0.1m) (Fig \ref{fig:fdm_evaluation}.A). Furthermore, we could simulate 1,500 of 6-second trajectories in about \unit[3]{ms} window with only using real-time exteroceptive sensor data, which is more than 20,000 times and 20 times faster than \textit{\complete} and \textit{\approximate} that use accurately constructed environment models (Fig \ref{fig:fdm_evaluation}.C), respectively. \textit{\complete} especially entailed very high computation costs. Thus, it was not suitable for applications that require lots of command samples to be simulated in real-time, such as our proposed algorithm. \textit{\approximate}, which simplifies the kinematic and dynamic model, could be a solution to alleviate the computation cost. However, such a method showed a large coordinate prediction error as the horizon became longer due to error accumulation (Fig \ref{fig:fdm_evaluation}.B). Our ablation study between \fdm and \textit{\approximate} will further explain the importance of prediction accuracy of forward simulation for the overall navigation performance.

In the following experiments, we will show how the effect of slight errors in \fdm can be mitigated by its low computational cost and other elements in the proposed framework for safe navigation. When using \fdm for safe navigation, we added a padding step. If \fdm predicted the first collision in $h$ step, the subsequent predicted outputs from $h$ step were set constant as $\textbf{x}_{t+h}=\textbf{x}_{t+h+1}..=\textbf{x}_{t+H}$, $\textbf{y}_{t+h}=\textbf{y}_{t+h+1}..=\textbf{y}_{t+H}$, and $\textbf{p}_{t+h}=\textbf{p}_{t+h+1}..=\textbf{p}_{t+H}$ $(1 \leq h \leq H)$.

\subsection{Point-Goal Navigation} \label{pgn_section}

In \textit{point-goal navigation} task, the robot should safely track the path, planned from the global planner, to navigate from the start location to the goal location. We compared the proposed method with a waypoint-based method that generates command using a PD controller, which will be referred to as \pd hereinafter. The detailed mechanism is explained in section \ref{related_work}. The desired orientation, used in \pd, was computed for the robot to be aligned with the planned path.

Three evaluation metrics were used to compare the performance: Success Rate (SR), Traversal Time, and Dynamic Time Warping (\dtw). Success here means that the robot reached the location within \unit[0.6]{m} from the goal without any collision with obstacles. \dtw was computed between the globally planned path and the robot COM's traversal path. We use \dtw per step to make the metric invariant to the path length.

We evaluated the performance in 60 different randomly generated open field environments with densely placed obstacles and 8 goal positions. As the proposed method relies on a random command sampler, we repeated the experiments with three different random seeds and the mean values of the evaluation metrics are reported in Table \ref{tab:pointgoalnav_result}. The standard deviations were small enough to be ignored.

\begin{table*}[t]
    \centering
    \begin{tabular}{l|c|c|c|c|c|c|c|c|c|c|c|c}
        Obstacle density [1/m] & \multicolumn{3}{c|}{0.43} & \multicolumn{3}{c|}{0.33} & \multicolumn{3}{c|}{0.25} & \multicolumn{3}{c}{0.2}  \\
        \hline & SR & Time & DTW & SR & Time & DTW & SR & Time & DTW & SR & Time & DTW \\
        \hline \hline \textbf{Ours} & \textbf{83.2} & 48.0 & 0.43 & \textbf{95.9} & 33.5 & 0.34 & \textbf{96.9} & 31.6 & 0.34 & 98.2 & 30.9 & 0.34 \\
        \hline \pd & 45.2 & 32.9 & 0.30 & 84.5 & 32.7 & 0.28 & 91.7 & 32.5 & 0.27 & 94.6 & 32.4 & 0.26 \\
        \hline Approximate \textit{(ABL)} & 76.5 & 43.8 & 0.44 & 93.4 & 33.2 & 0.36 & 95.9 & 31.9 & 0.35 & 98.3 & 31.4 & 0.35 \\
        \hline only Random \textit{(ABL)} & 73.6 & 51.7 & 0.45 & 89.5 & 34.2 & 0.35 & 96.6 & 31.8 & 0.34 & \textbf{98.6} & 31.0 & 0.34 \\
        \hline only \its \textit{(ABL)} & 63.2 & 39.3 & 0.38 & 90.3 & 33.0 & 0.35 & 91.5 & 31.6 & 0.36 & 91.9 & 31.1 & 0.37 \\
    \end{tabular}
    \caption{\textit{Point-Goal Navigation} results in open fields with densely placed obstacles. Success rate (SR) [\%], traversal time [s], and \dtw [m] are reported. The results are the mean values after running the evaluation with three different random seeds. Obstacle density represents the number of obstacles per meter and is computed as $\frac{1}{obstacle\ grid\ size}$. Candidates with a \textit{ABL} flag were evaluated for the ablation study.}
    \label{tab:pointgoalnav_result}
    \squeezeup
\end{table*}

Our method showed a higher success rate compared to \pd with similar traversal time and \dtw. \pd suffered from collisions in complex and narrow environments (i.e. environments with high obstacle density) due to unaccounted safety during the local planning. Performance of \its will be further analyzed in section \ref{ablation}.

\begin{figure}[t]
    \centering
    \includegraphics[width=\linewidth]{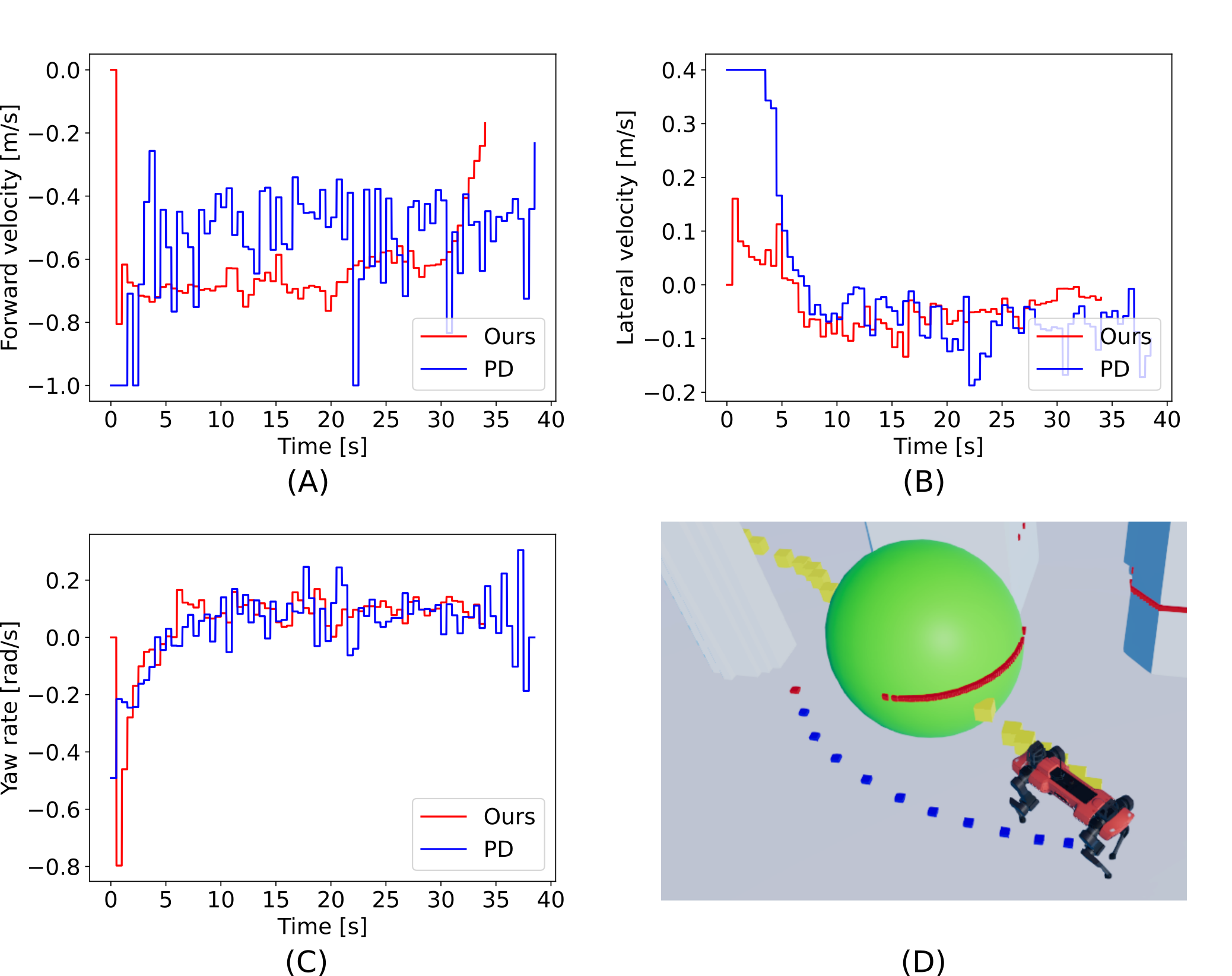}
    \caption{(A-C) shows the velocity command trajectory executed on the robot to reach the goal. (A), (B), (C) each corresponds to forward velocity, lateral velocity, yaw rate command. (D) shows the reactiveness of our method to avoid unexpected obstacles (green) located on the globally planned path.}
    \label{fig:command_and_reactiveness}
\end{figure}

Because the proposed model-predictive control module iteratively replans a long horizon command trajectory to improve both path similarity and safety, our local planner resulted in a smoother command trajectory on the robot than \pd (Fig. \ref{fig:command_and_reactiveness}.A-C). Furthermore, it could handle unexpected obstacles on the globally planned path and avoid them without a new plan from the global planner (Fig. \ref{fig:command_and_reactiveness}.D).

\begin{figure}[t]
    \centering
    \includegraphics[width=\linewidth]{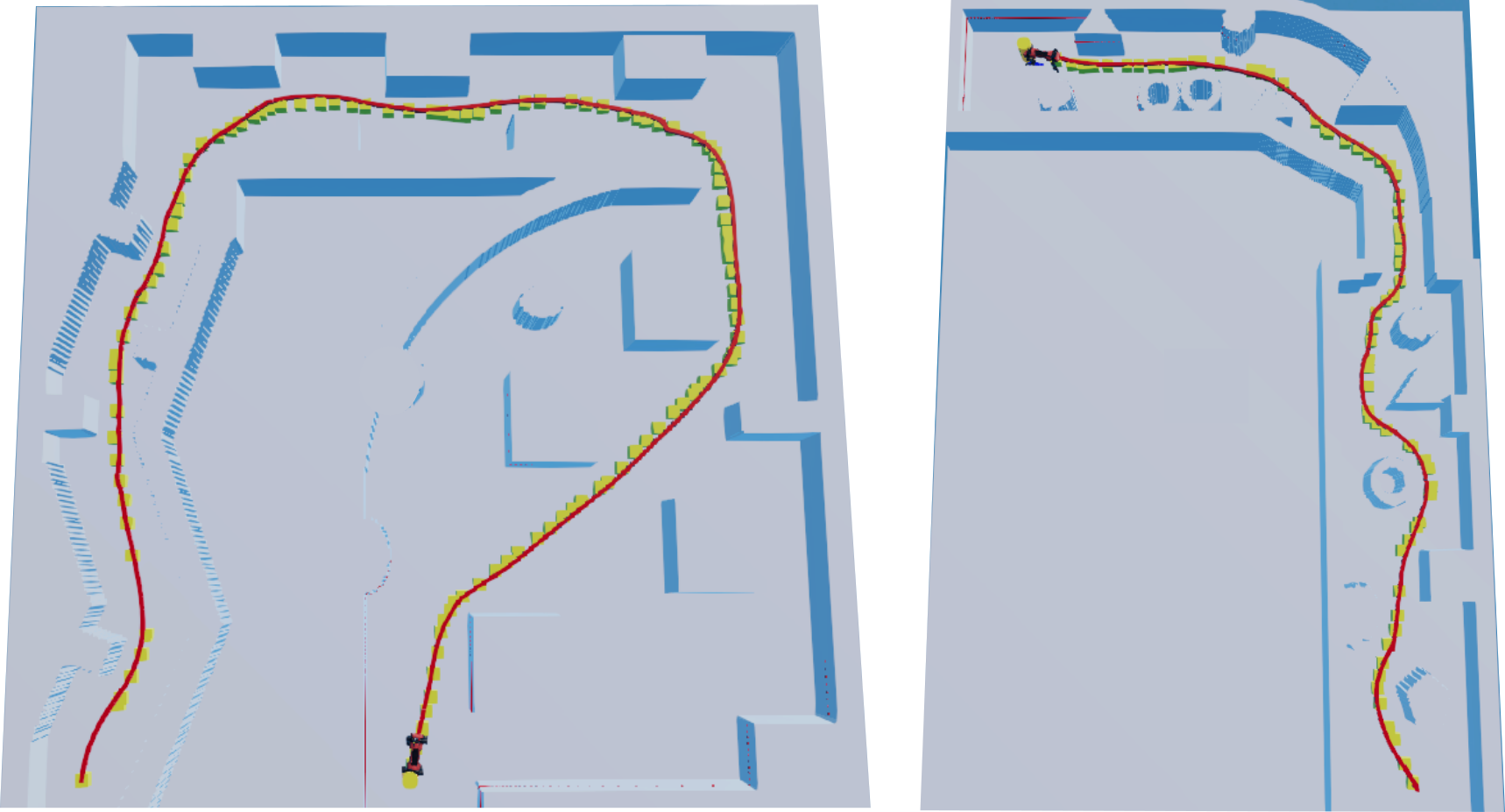}
    \caption{Map and the robot's traversal path in novel environments}
    \label{fig:pgn_novel}
    \squeezeup
\end{figure}

To check the generalizability and robustness of our method, we generated two different environments with various geometric complexity as shown in Fig. \ref{fig:pgn_novel}. Visualized traversal paths of the robot in each environment indicate that our method could safely navigate over a long distance.

\subsection{Semi-autonomous task}

\begin{figure}[t]
    \centering
    \includegraphics[width=\linewidth]{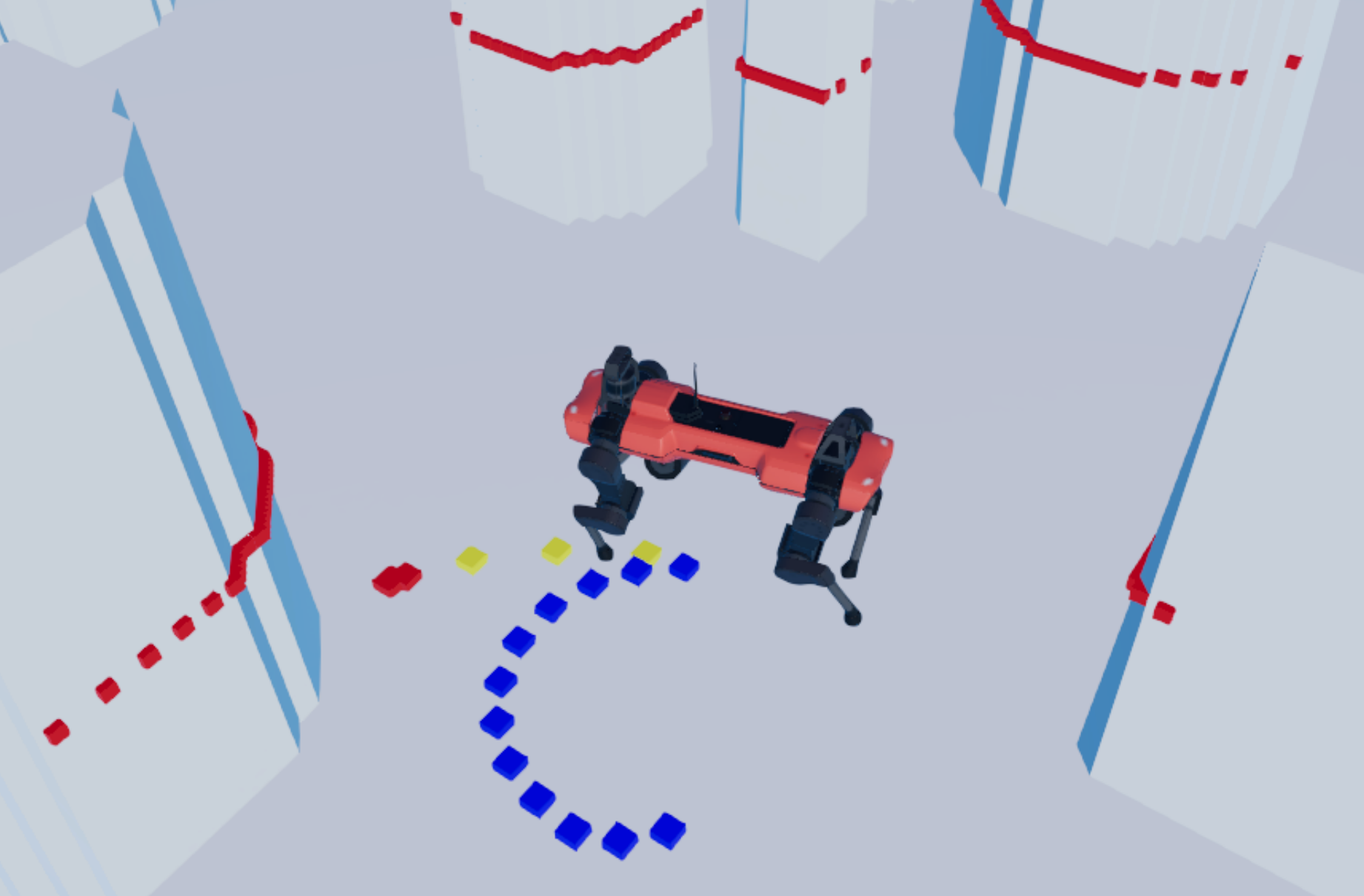}
    \caption{Visualization of \textit{Safety Remote Control} task. The blue dotted line is the generated trajectory by the optimized command. The yellow and red dotted line is a generated trajectory by the given command. The red dots indicate that \fdm predicts collision along the trajectory.}
    \label{fig:src}
\end{figure}

\begin{table}[t]
    \centering
    \begin{tabular}{l|c|c|c|c}
        Obstacle density [1/m] & 0.4 & 0.33 & 0.25 & 0.2 \\
        \hline \hline Collision safe [\%] & 81.0 & 85.8 & 88.4 & 88.4 \\
        \hline No collision safe [\%] & 98.2 & 98.8 & 99.8 & 99.8 \\
    \end{tabular} 
    \caption{\textit{Safety Remote Control} result in open fields with densely placed obstacles. Success rate (SR) [\%] is reported. Obstacle density represents the number of obstacles per meter and is computed as $\frac{1}{obstacle\ grid\ size}$. “Collision safe” and “No collision safe” each represent the success case when directly executing the given command results in a collision or not. To be specific, “Collision safe” corresponds to “(number of collision-free trajectories among the trajectories when directly executing the given command results in a collision) / (number of trajectories when directly executing the given command results in a collision)”. “No collision safe” corresponds to “(number of collision-free trajectories among the trajectories when directly executing the given command does not result in a collision) / (number of trajectories when directly executing the given command does not result in a collision)”}
    \label{tab:safetyremotecontrol_result}
    \squeezeup
\end{table}

\fdm and model-predictive control module with a random sampler are not limited to fully-autonomous tasks like \textit{point-goal navigation} and can also be applied to semi-autonomous tasks. We evaluated our proposed method on a semi-autonomous task named \textit{safety remote control}. In \textit{safety remote control} task, the robot should decide whether the given remote command from the user is safe and project the command onto the safe command set if it is not. (Fig. \ref{fig:src}).

We used $R_{safety}$ and the proposed model-predictive control algorithm for the task with an additional simple logic. If \fdm predicted the given user command to collide with obstacles within a fixed period (3s), it executes the optimized command. On the other hand, if the given command was predicted as safe, it executes the given command. We sampled the commands for optimization in a multivariate normal distribution with given user command as mean values to search solutions close to the user's intended command.

We evaluated the performance in open field environments with densely placed obstacles. In each environment, 300 randomly sampled commands were given to the robot and recorded as a success if the robot did not make a contact with the environment other than on the feet, which is the same rule applied for \textit{point-goal navigation}. Results shown in Table \ref{tab:safetyremotecontrol_result} indicate that our method with \fdm can predict collision and project the command onto the safe command set, if needed, with a high accuracy. Furthermore, it shows that our proposed method can be used for both fully-autonomous and semi-autonomous tasks by just changing the command sampling distribution and reward functions.

\subsection{Ablation study} \label{ablation}

In the ablation study, we aim at showing the importance of the following contributions for the overall performance in safe navigation:
\begin{itemize}
    \item Learned \fdm that outputs the future coordinates and the collision probabilities
    \item Usage of both random sampler and learned \its for command sampling
\end{itemize}
For evaluation, we used the same evaluation metrics and procedure explained in section \ref{pgn_section}.

First, we used \textit{Approximate} described in section \ref{fdm_evaluation}, instead of \fdm, for the proposed navigation framework and evaluated the performance (Table \ref{tab:pointgoalnav_result}). \textit{Complete} was not used because of the heavy computation for real-time usage. Ours using \fdm showed higher success rate than ones with \textit{Approximate}. This was due to the large coordinate prediction error shown in \textit{Approximate} (Fig \ref{fig:fdm_evaluation}.B). It caused the predicted trajectory to deviate a lot from the actual trajectory and thus returned an inaccurate reward signal for the optimization. Furthermore, compared to the physics simulator that outputs discrete collision state, \fdm outputs continuous collision probability which enabled distinct safety reward signal for each sample.

\begin{figure}[t]
    \centering
    \includegraphics[width=\linewidth]{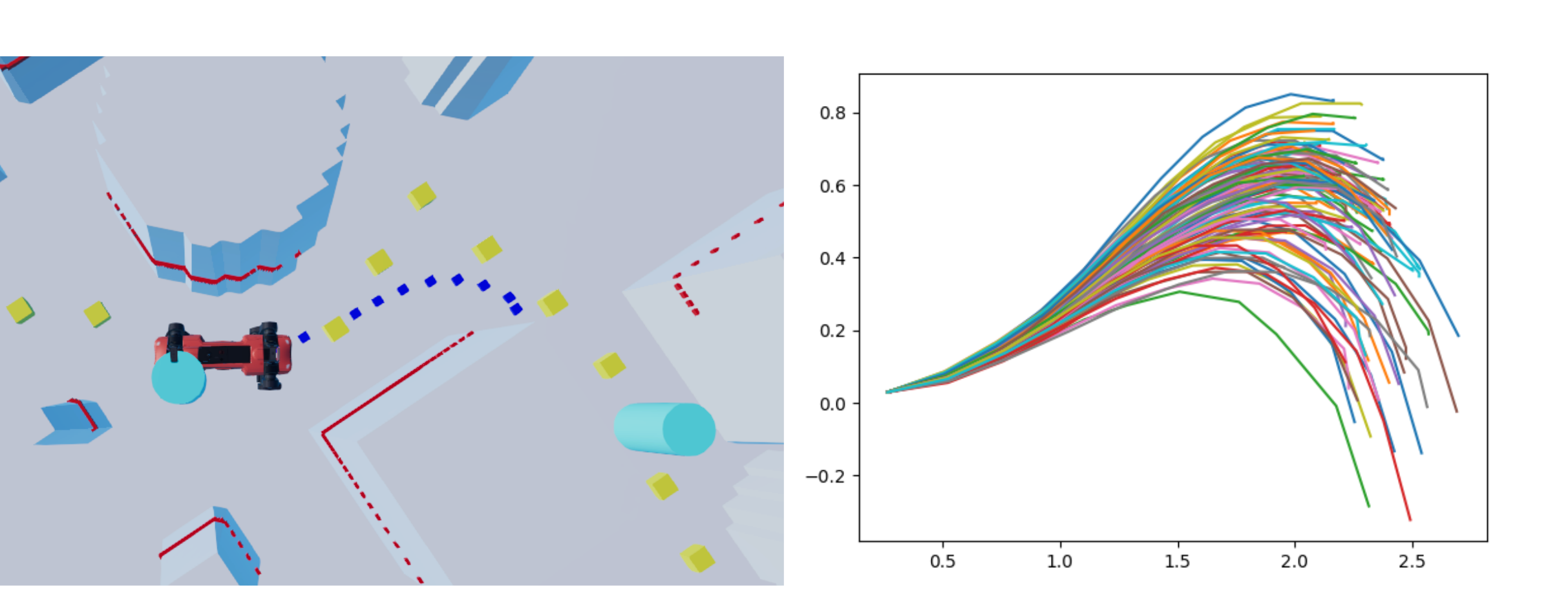}
    \caption{Visualization of trajectories sampled using \its (Right) and the output trajectory from the optimizer using these samples (Left). The robot is placed at the origin. The yellow dotted line is the global path to track. The yellow dotted line between two light-blue cylinders is the waypoint trajectory considered when generating the visualized samples. As \its samples command sequences, outputs of learned \fdm is used to visualize the trajectories.}
    \label{fig:its_sampled_traj}
    \squeezeup
\end{figure}

To check the importance of the command sampler, ours were compared with two methods, each using only the random sampler or \its. Results in Table \ref{tab:pointgoalnav_result} indicate the complementary relationship between random sampler and \its for safe navigation. The random sampler generates diverse command sequences and enables the optimizer to find an approximate solution. However, it fails to find the precise solution which is critical in complex and narrow environments. On the other hand, \its can guide the optimizer to find a more precise solution by generating biased command sequences (Fig. \ref{fig:its_sampled_traj}). However, it fails to generate diverse samples, compared to the random sampler, to handle a local optimum and thus showed relatively low performance when used alone.


\section{Conclusion}

We proposed a learning-based fully autonomous navigation framework composed of three innovative elements: a learned forward dynamics model (\fdm), an online sampling-based model-predictive control module, and an informed trajectory sampler (\its). We demonstrated how the dynamics of a quadruped robot and its surroundings can be learned accurately using deep neural networks (\fdm) and be used for various downstream tasks, including safe \textit{point-goal navigation}, with the sampling-based model-predictive control module. To handle the curse of dimensionality in sampling-based motion planning, we further suggested the informed sampler, composed of deep neural networks, and its training method. Extensive evaluation in the physics simulator~\cite{hwangbo2018per} showed the superiority of the performance of the proposed method for autonomous navigation in complex environments compared to the widely used baseline method. The ablation studies further demonstrated how the elements of our framework worked in a complementary manner resulting in a high-performance navigation system.

Promising directions for future work include expanding our method to different robot platforms (e.g. drone, wheeled robot, biped robot) using different exteroceptive sensors (e.g. RGB/D camera, 3D lidar sensor) and transferring it to the real world.


\section*{ACKNOWLEDGMENT}
This work was supported by Samsung Research Funding \& Incubation Center of Samsung Electronics under Project Number SRFC-IT2002-02.

\bibliographystyle{plainnat}
\bibliography{main}

\end{document}